\documentclass[letterpaper]{article} 
\usepackage{aaai23}  
\usepackage{times}  
\usepackage{helvet}  
\usepackage{courier}  
\usepackage[hyphens]{url}  
\usepackage{graphicx} 
\usepackage{natbib}  
\usepackage{caption} 
\usepackage{algorithm}
\usepackage{algorithmic}

\usepackage{newfloat}
\usepackage{listings}
\DeclareCaptionStyle{ruled}{labelfont=normalfont,labelsep=colon,strut=off} 
\floatstyle{ruled}
\newfloat{listing}{tb}{lst}{}
\floatname{listing}{Listing}

\usepackage{times}
\usepackage{soul}
\usepackage{url}
\usepackage[hidelinks]{hyperref}
\usepackage[utf8]{inputenc}
\usepackage{amsmath}
\usepackage{booktabs}
\usepackage{algorithm}
\usepackage{algorithmic}
\usepackage{lineno}
\usepackage{bm}
\usepackage{bbm}
\usepackage{float}
\usepackage{multirow}
\usepackage{makecell}
\usepackage{tabularx}
\usepackage{amssymb}
\usepackage{stmaryrd}
\usepackage{lipsum}
\usepackage{subfigure}
\usepackage{bbold}

\newtheorem{definition}{Definition}

\newtheorem{theorem}{Theorem}

\title{Robust Causal Graph Representation Learning against Confounding Effects}

\author{
    Hang Gao\textsuperscript{\rm 1}\textsuperscript{\rm 2}\equalcontrib, Jiangmeng Li\textsuperscript{\rm 1}\textsuperscript{\rm 2}\equalcontrib\thanks{Corresponding author.}, Wenwen Qiang\textsuperscript{\rm 1}\textsuperscript{\rm 2}, Lingyu Si\textsuperscript{\rm 1}\textsuperscript{\rm 2}, Bing Xu\textsuperscript{\rm 3}, \\Changwen Zheng\textsuperscript{\rm 1}, Fuchun Sun\textsuperscript{\rm 4}
}
\affiliations{
    \textsuperscript{\rm 1}Science and Technology on Integrated Information System Laboratory, \\Institute of Software Chinese Academy of Sciences\\
    \textsuperscript{\rm 2}University of Chinese Academy of Sciences\\
    \textsuperscript{\rm 3}China Communications Technology Information Group Co., Ltd.\\
    \textsuperscript{\rm 4}Tsinghua University\\
    \{gaohang, jiangmeng2019, qiangwenwen, lingyu, changwen\}@iscas.ac.cn,\\ xubing@ccccltd.cn, fcsun@mail.tsinghua.edu.cn
}

\usepackage{bibentry}

\begin{document}

\maketitle

\begin{abstract}
The prevailing graph neural network models have achieved significant progress in graph representation learning. However, in this paper, we uncover an ever-overlooked phenomenon: the pre-trained graph representation learning model tested with full graphs \textit{underperforms} the model tested with well-pruned graphs. This observation reveals that there exist confounders in graphs, which may interfere with the model learning semantic information, and current graph representation learning methods have not eliminated their influence. To tackle this issue, we propose \textit{\textbf{R}obust \textbf{C}ausal \textbf{G}raph \textbf{R}epresentation \textbf{L}earning} (RCGRL) to learn robust graph representations against confounding effects. RCGRL introduces an \textit{active} approach to generate instrumental variables under unconditional moment restrictions, which empowers the graph representation learning model to eliminate confounders, thereby capturing discriminative information that is causally related to downstream predictions. We offer theorems and proofs to guarantee the theoretical effectiveness of the proposed approach. Empirically, we conduct extensive experiments on a synthetic dataset and multiple benchmark datasets. The results demonstrate that compared with state-of-the-art methods, RCGRL achieves better prediction performance and generalization ability. Our codes are available at https://github.com/hang53/RCGRL.
\end{abstract}


\section{Introduction}
``Graph'' is a derived discrete data structure consisting of vertices and edges, which can be leveraged to model and solve various general problems. Benefitting from the embedding of human knowledge, graphs are semantically dense data. In contrast, native data structures, such as images and videos, are usually semantically sparse data. Therefore, effectively using graphs to model and learn valuable information for downstream tasks is a compelling area of research. The impressive success in Graph Neural Networks (GNNs) \cite{kipf2016semi, xu2018powerful, velivckovic2017graph} provokes the exploration to sufficiently learn \textit{discriminative} representations from graphs.

\begin{figure}[t]
    \centering
    \includegraphics[width=0.4\textwidth]{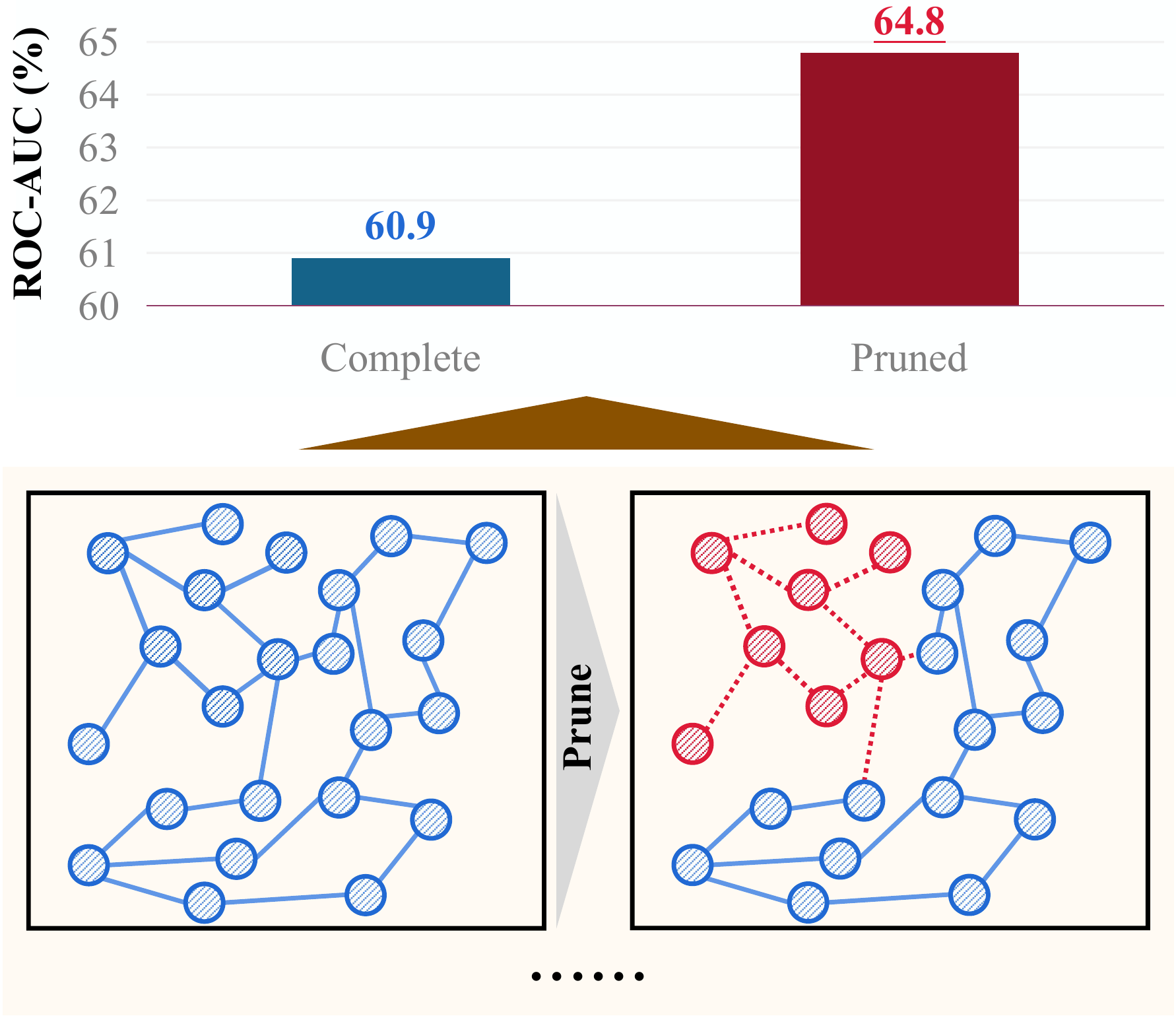}
    \caption{Comparisons achieved by the baseline, i.e., Empirical Risk Minimization (ERM). We fix the GNN model trained on Mol-BBBP \cite{hu2020open} and then randomly select a batch of data from the test set of Mol-BBBP. The results of \textit{Complete} and \textit{Pruned} variants are obtained by ERM on complete and pruned test graphs, respectively.}
    \label{fig:motiv}
\end{figure}

To this end, we revisit the learning paradigm of state-of-the-art methods and conclude that the improvements according to GNNs mainly include 1) the message passing process, which is the operation to transform and aggregate features from a node's graph neighborhood, e.g., Graph Convolution Networks \cite{kipf2016semi}, Graph Isomorphism Networks \cite{xu2018powerful}, Graph Attention Networks \cite{velivckovic2017graph}, and so on; 2) the readout function, which is the process to pool the node representations into the graph representation, e.g., Adaptive Structure Aware Pooling \cite{DBLP:conf/aaai/RanjanST20}, Top-k Pool \cite{DBLP:conf/icml/GaoJ19}, SAG Pool \cite{DBLP:conf/icml/LeeLK19}, etc. Benchmark methods focus on learning semantic information from \textit{complete} graphs by proposing elaborate GNN structures. However, are all substructures in a graph valuable for predicting downstream tasks? We borrow benchmark learning paradigms from other fields. In the field of visual classification, object detection, as a typical visual task, requires the model to detect the \textit{foreground} by sufficiently excluding the \textit{background} objects and then performing classification for each foreground object. The background objects mentioned here can have a confounding effect on the model's predictions. Inspired by this paradigm, we intuitively propose that, in graph representation learning, some parts of the data also have confounding effects that need to be eliminated.


\begin{figure}[t]
    \centering
    \includegraphics[width=0.4\textwidth]{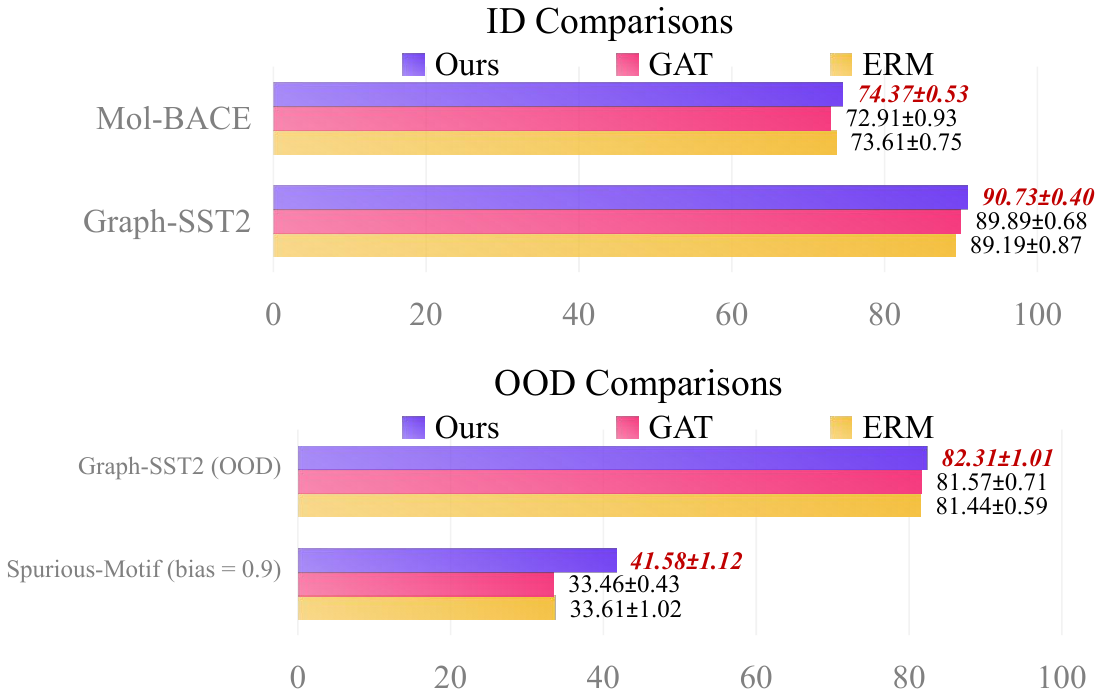}
    \caption{Comparisons achieved by the baselines, i.e., ERM and Graph Attention Network (GAT) \cite{velivckovic2017graph}. Graph-SST2 \cite{yuan2020explainability} and Mol-BACE \cite{hu2020open} are the benchmark in-domain (ID) datasets. To evaluate the generalization ability of GNN models, we build out-of-domain (OOD) datasets by following the principle of DIR \cite{wu2022discovering}. Specifically, Spurious-Motif is an artificially generated dataset. Each graph in Spurious-Motif consists of two subgraphs, a ground-truth subgraph, and a confounder subgraph. In the training set, these subgraphs are patched together under certain rules. In contrast, in the validation and test sets, these subgraphs are patched together randomly. The term \textit{bias} measures the difference between the training set and the other sets. Graph-SST2 (OOD) is a manually regrouped dataset, where the graphs are split into different sets according to the average node degrees.}
    \label{fig:motividood}
\end{figure}

To demonstrate our statement, we run some motivating experiments. In Figure \ref{fig:motiv}, we iteratively remove specific substructures from the original graph to generate multiple pruned graphs by following the first principle of Graph Granger Causality \cite{DBLP:conf/icml/LinLL21}. We prune the substructures of graphs that interfere with the predictions. In this work, we name such substructures as \textit{graph confounders}. The empirical results demonstrate that compared with the original graphs, the pre-trained GNN yields performance boosts on graphs pruned by the well-selected scheme. Such results prove the existence of the proposed graph confounders, and such confounders can indeed interfere with the downstream predictions. Note that the pruning requires the label information of the test data. Thus, this motivating experiment can only be introduced to prove the existence of graph confounders. From the experiments in Figure \ref{fig:motividood}, we observe that our method, learning robust graph representation against confounding effects, generally outperforms benchmark methods that learn semantic information from complete graphs (containing confounders) on both ID and OOD tasks by significant margins. These empirical results demonstrate that graph confounders may degenerate the prediction performance and generalization ability of GNNs. The reason behind such observations is that graph confounders impede the model's acquisition of discriminative information that is causally related to downstream predictions. Therefore, the prediction performance of learned representations degenerates. Furthermore, due to the confounding effects, the learned representations may contain data-biased information so that the generalization ability of the learned representations is degraded.

To learn graph representations against confounding effects, we propose a novel approach, called \textit{\textbf{R}obust \textbf{C}ausal \textbf{G}raph \textbf{R}epresentation \textbf{L}earning} (RCGRL), which aims to obtain the graph representations that is causally related to the prediction on downstream tasks. To this end, RCGRL adopts the concept of the Instrumental Variable (IV) approach \cite{wooldridge2002econometric}. IV approach defines the conditional moment restrictions under which the confounding effect could be eliminated. Accordingly, the approach searches for suitable IVs so that the model can satisfy the mentioned restrictions. However, due to the complicity of graph representation learning models, the conditional moment restrictions are hard to satisfy, so the corresponding IVs are difficult to obtain. Therefore, our implementation of the IV approach contains a novel method that will actively generate the expected IVs, thereby transferring the conditional moment restrictions into unconditional ones. We provide theoretical proof to demonstrate that our implementation achieves the same effect as the IV approach under conditional moment constraints in Theorem \ref{the:ivrules} and the corresponding proof for Theorem \ref{the:ivrules}. Theorem \ref{the:ivrules} also ensures the objective of such a GNN leading the network to generate desired IVs. Guided by the theorem, we design two novel training objectives, i.e., the robustness-emphasizing loss and the contrast-auxiliary loss, to empower the model to learn robust graph representations against confounding effects. The graph representation learning network and the IV generating network are updated alternatively until convergence. Empirically, we evaluate our proposed RCGRL on one synthetic and three real datasets. Extensive experiments demonstrate that the classification performance on ID tasks and the generalization ability on OOD tasks of RCGRL surpass state-of-the-art graph representation learning methods. The following are our contributions:

\begin{itemize}
\item We propose a novel method, RCGRL, to learn robust graph representations against confounding effects, thereby improving the prediction performance and generalization ability of the representations. Such an approach is suitable for general deep graph neural network models.

\item Theoretically, we offer proof to demonstrate the performance guarantee of implementing the IV approach under unconditional moment constraints and a theorem to prove the effectiveness of the generated IVs.

\item We provide the implementations of RCGRL for prediction and generalization tasks. The results demonstrate the consistently excellent performance of RCGRL on a synthetic dataset and three benchmark datasets.

\end{itemize}

\section{Related Work}

\subsection{Graph Neural Networks}
GNN learns the representations of graphs by combining neural networks with graph learning. Such representations can be applied to a variety of downstream tasks. Like other neural network structures, GNNs developed multiple variants. Graph Convolution Network (GCN) \cite{kipf2016semi} conducts graph learning utilizing the convolutional network. \cite{xu2018powerful} propose Graph Isomorphism Network (GIN) which possesses the same representation capability as the Weisfeiler-Lehman test. Graph Attention Network (GAT) \cite{velivckovic2017graph} adopts the concept of attention mechanism and enables the network to focus on certain parts of the graph data. To improve the pooling process of GNNs, Graph-u nets \cite{gao2019graph} propose a novel GNN with an adaptive pooling mechanism.


\subsection{Graph Causality Learning}
Causality learning identifies causal relationships between data and labels. In economics and statistics, means of identifying causality have been studied for a long time, and various methods have been proposed \cite{glymour2016causal, wooldridge2002econometric, li2014u}. Recently, the methods of causality learning have also been adopted in deep learning to optimize the rationality and robustness of the model. In graph learning, studying the causality within models is also beneficial. \cite{ying2019gnnexplainer, luo2020parameterized, yuan2021explainability} propose to adopt an explanation method to figure out the causal relationship between the model's inputs and outputs. \cite{DBLP:conf/icml/LeeLK19, velivckovic2017graph} create intrinsic interpretable learning frameworks by incorporate rationalization modules, including graph attention and pooling. \cite{wu2022discovering} introduces the intervention operation from the causal inference theory \cite{glymour2016causal} and builds causal models through finding generalization causally related information. Our method, on the other hand, improves the causality by learning robust graph representations against confounding effects.

\section{Method}

\subsection{Preliminary}
First, we briefly review some necessary preliminary concepts and notations. 

\subsubsection{IV Approach.}
IV approach \cite{wooldridge2002econometric} deals with the confounding effects by conducting estimations under conditional moment restrictions. The general type of an IV model can be described as follows. Let $X_{i}$, $Y_{i}$ denote observable variable vectors of some data distribution to be studied. $X_{i}\in \mathcal{R}^{m}$, $Y_{i}\in \mathcal{R}^{b}$. Assume that the observations $\{(X_{i}, Y_{i})\}^{n}_{i=1}$ are i.i.d., our goal is to learn a structural function $f\left(\cdot\right)$ that can represent the correlations between $X_{i}$ and $Y_{i}$. We presume that $f\left(\cdot\right)$ can be completely determined by parameters $\theta^{*}$. Such a function can be formulated as:
\begin{align}
	Y_{i} = f_{\theta^{*}}(X_{i}).
	\label{eq:gnn}
\end{align}
Then, we can acquire $\theta^{*}$ using regression methods. Unfortunately, in real-world circumstances, confounders generally exist in the data. Formally, we have:
\begin{align}
	Y_{i} =  f_{\theta^{*}}(X_{i}) + \epsilon_{i},
	\label{eq:gnn}
\end{align}
where $\epsilon_{i}$ is the error term. If we cannot acquire a clear representation of $\epsilon_{i}$, we will not be able to get ${\theta^{*}}$ correctly according to $Y_{i}$ and $X_{i}$. 

The IV approach addresses this issue with conditional moment restrictions. Such an approach selects a conditioning variables vector $Z_{i}$ as the IV. $\{(X_{i}, Y_{i},Z_{i})\}^{n}_{i=1} \text{ are i.i.d.}$, $Z_{i}$ satisfies:

\begin{align}
    \mathbb{E}[\epsilon_{i}|Z_{i}]=0 \ \ \  \forall i \in \llbracket 1, n \rrbracket .
	\label{eq:gnn}
\end{align}

By introducing $Z_{i}$ into the calculation of $f\left(\cdot\right)$, we can eliminate the confounding effects brought by $\epsilon_{i}$. Based on this idea, \cite{ai2003efficient} defines a formal training objective of the IV approach as:
\begin{align}
	 \mathbb{E}[f_{\theta^{*}}(X_{i}) - Y_{i} | Z_{i}]=0.
	\label{eq:fx}
\end{align}
 However, the IV, $Z_{i}$, is also hard to acquire in most learning tasks. In linear cases, approaches, such as \cite{hayashi2000econometrics, hansen1996finite}, address the issue by transferring the conditional moment restrictions into unconditional ones. 
 
 \subsubsection{Graph Representation Learning.}

Given an attributed graph $G = (V, E)$, $V$ and $E$ denote the node and edge sets, respectively. $\mathcal{G} = \left\{G_i, i \in \llbracket{1, N^G} \rrbracket\right\}$ is a graph dataset. The objective of graph representation learning is to learn an encoder $f^G(\cdot): \mathcal{G} \to \mathbb{R}$, where $\mathbb{R}$ denotes an embedding space. Accordingly, $f^G(G_i)$ is the representation containing discriminative information of $G_i$ for downstream tasks.

To learn the discriminative representation, benchmark methods employ GNN as the encoder. GNN encodes nodes in $G_i = (V_i, E_i)$ into embedded vectors. The $k$-th layer of GNN can be formulated as:
\begin{equation}
	\boldsymbol{h}^{\left(k+1\right)}_{v} = \textrm{COM}^{\left(k\right)} \Bigg(\boldsymbol{h}^{\left(k\right)}_{v}, \textrm{AGG}^{\left(k\right)} \left(\boldsymbol{h}^{\left(k\right)}_{u} ,\forall u \in \textrm{N}\left(v\right)\right)\Bigg),
	\label{eq:gnnsub}
\end{equation}
where $\boldsymbol{h}_{v}$ denotes the vector for node $v \in V_i$, $\textrm{N}\left(v\right)$ denotes the neighbors of $v$, and $\boldsymbol{h}^{\left(k\right)}$ is the representation vector of the corresponding node at the $k$-th layer. Specifically, $\boldsymbol{h}^{\left(0\right)}$ is initialized with the input node features. $\textrm{COM}\left(\cdot\right)$ and $\textrm{AGG}\left(\cdot\right)$ are learnable functions of GNN. $\textrm{COM}\left(\cdot\right)$ combines the aggregated neighbor feature into the feature of the target node, and $\textrm{AGG}\left(\cdot\right)$ aggregates the features of neighbors. To obtain the graph representation $f^G(G_i)$ for $G_i$, we adopt
\begin{equation}
	{f^G(G_i)} = \textrm{READOUT} \left(\boldsymbol{h}_v, v \in V_i\right),
	\label{eq:gnn}
\end{equation}
where $\textrm{READOUT}\left(\cdot\right)$ is a readout function pooling the node representations $\left\{ \boldsymbol{h}_{v} | v\in V_i \right\}$.
 
\begin{figure*}[t]
\centering
\includegraphics[width=0.9\textwidth]{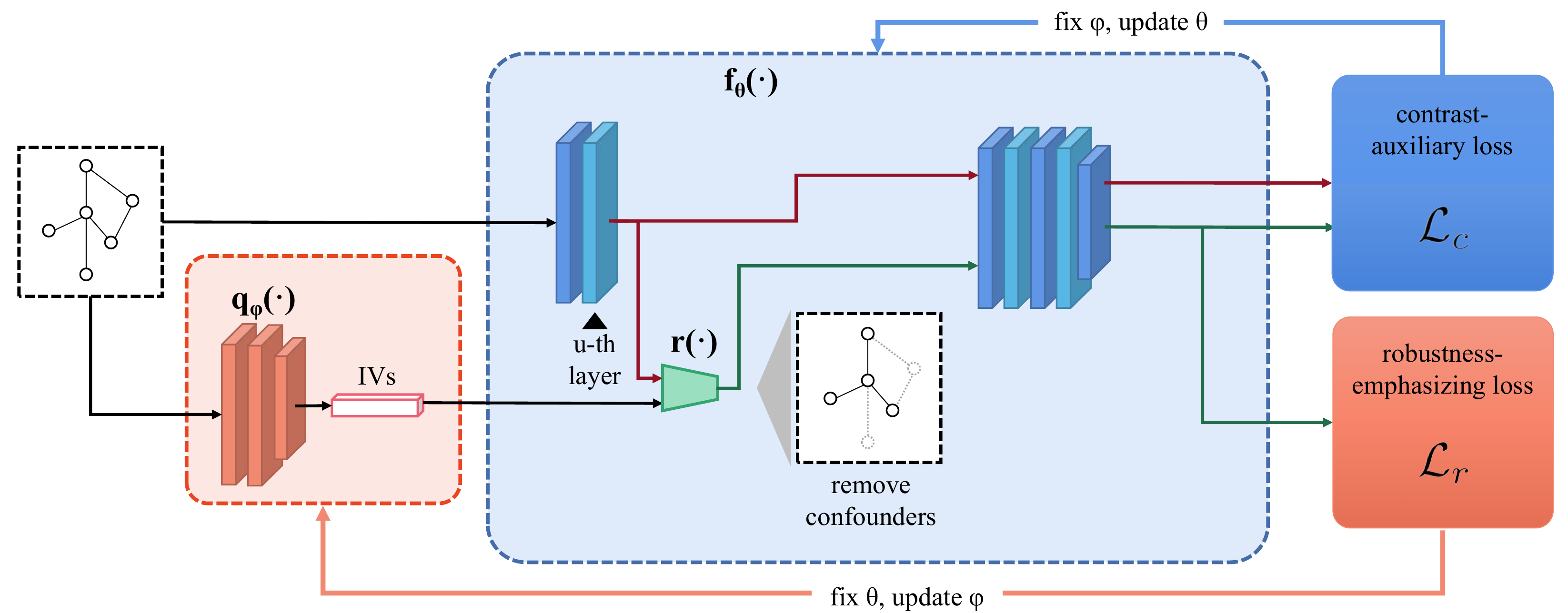} 
\caption{RCGRL's framework. In the figure, the blue neural network represents $f(\cdot)$, and the orange-red represents $q(\cdot)$.}
\label{fig:framework}
\end{figure*}
 
\subsection{Robust Causal Graph Representation Learning}
 
 
\subsubsection{Theoretical Design of RCGRL.}
Our method aims at enhancing causality in graph representation learning. We propose that models generating predictions with high causal relationships with labels can be treated as ``confounder-robust'' models. We define a confounder-robust GNN as follows.
 
\begin{definition} 
(Confounder-robust GNN). Given a graph dataset $\mathcal{G}$, if a GNN $f(\cdot)$ outputs information that contains the sufficient and correct information of $\mathcal{G}$ to perform prediction in downstream tasks, regardless of confounder $C$, then $f(\cdot)$ is a confounder-robust model in $\mathcal{G}$.
\label{def:aug}
\end{definition} 
 
 We adopt the IV approach~\cite{greene2003econometric} to empower us to acquire a confounder-robust model. Given a graph dataset $\mathcal{G} = \left\{G_i, i \in \llbracket{1, N^G} \rrbracket\right\}$, we define the confounder in each graph as $C_{i}$. Meanwhile, $X_{i}$ denotes the information causally related to $Y_{i}$. However, $X_{i}$ and $C_{i}$ are mixed in $G_{i}$. Our purpose is to train a GNN model to predict the label $Y_{i}$ based on $X_{i}$, regardless of $C_{i}$. We denote the parameter of the model that can achieve such a goal as $\theta^{*}$. With the IV approach, our objective is to acquire $\theta^{*}$, which can be formulated as follows.
\begin{gather}
	 \theta^{*} = \arg\min_{\theta}\Big(\mathbb{E}[m(G_{i}, Y_{i}, Z_{i},  \theta)|Z_{i}]\Big), \forall i \in \llbracket 1, n \rrbracket \nonumber\\s.t. \ \mathbb{E}[(|C_{i}|)|Z_{i}]=0,  \  \forall i \in \llbracket 1, n \rrbracket ,
	\label{eq:goal1}
\end{gather}
where $m(\cdot)$ outputs the difference between the model's output and ground truth labels, $|C_{i}|$ denotes the number of elements in $C_{i}$, $Z_{i}$ denote the IVs that instruct the model to eliminate confounders $C_{i}$ through certain measures. Equation \ref{eq:goal1} can be treated as a generalization of Equation \ref{eq:fx} in the field of graph representation learning.

As mentioned above, the IV $Z_{i}$ can be challenging to acquire. Furthermore, the complexity of graph representation learning with neural networks enlarges such difficulty. To make calculating $Z_{i}$ operable, we transfer the conditional moment restrictions of Equation \ref{eq:goal1} into unconditional moment restrictions by generating the IV $Z_{i}$ with a GNN model $q^G(\cdot)$, which can be formulated as:
\begin{align}
	  Z_{i} = q^{G}_{\varphi}(G_{i}),
	\label{eq:z}
\end{align}
where $\varphi$ are the parameters of GNNs that generates $Z_{i}$. To replace $Z_{i}$, an ideal $q^{G}(\cdot)$ with parameters $\varphi^{*}$ should satisfy:
\begin{align}
\mathbb{E}[(|C_{i}|)|q^{G}_{\varphi^{*}}(G_{i})]=0, \ \ \  \forall i \in \llbracket 1, n \rrbracket.
	\label{eq:q1}
\end{align}
As we expect to train the model to predict the label $Y_{i}$ based on $X_{i}$, we need to leave $X_{i}$ unaffected. Therefore, $q^G(\cdot)$ should also satisfy:
\begin{align}
\mathbb{E}[X_{i}|q^{G}_{\varphi^{*}}(G_{i})]=X_{i}, \ \ \  \forall i \in \llbracket 1, n \rrbracket.
	\label{eq:q2}
\end{align}
As $q^G(\cdot)$ is a GNN, we can utilize the backpropagation to train it. We propose the following theorem to define the objective for training $q^G(\cdot)$.

\begin{theorem} \label{the:ivrules}
Given a large enough graph dataset $\mathcal{G}$, where each graph $G_{i} \in \mathcal{G}$ , $i \in \llbracket 1, n \rrbracket$. $G_{i}$ consists of two parts, $C_{i}$ and $X_{i}$. $C_{i}$ is the confounder. $X_{i}$ is causally related to the label $Y_{i}$. We define a model consisting of two GNNs, $f^G(\cdot)$ and $q^G(\cdot)$. $q^{G}(\cdot)$ outputs IVs that are used to remove the confounders in $G_{i}$. $f^{G}(\cdot)$ is a GNN that predicts $Y_{i}$. Assuming $\theta^*$ is sufficiently learned so that $f^{G}_{\theta^*}(X_{i})=Y_{i}$ and $f^{G}_{\theta^*}(X_{i}\cup C'_{i})\neq Y_{i}, \forall C'_{i} \subseteq C_{i}$. Then, we can acquire the parameters $\varphi^{*}$ that enable the $q^G(\cdot)$ to satisfy Equations \ref{eq:q1} and \ref{eq:q2} by maximizing the cross entropy between $Y_{i}$ and the model's output. Formally, we have:
\begin{align}
\varphi^{*} = \arg\max_{\varphi}\Big(I\left(f_{\theta^{*}}^G\left(r\left(G,q^{G}_{\varphi}(G)\right)\right),Y\right)\Big),
	\label{eq:fai}
\end{align}
where $I(\cdot)$ denotes the mutual information, $r(\cdot)$ represents a function without trainable parameters to remove confounders based on IVs.
\label{thm:main}
\end{theorem} 

\emph{Proof.} We provide proofs to demonstrate the effectiveness of Theorem \ref{the:ivrules}. Since we already know the values of $\theta^*$, therefore,  $\varphi^{*}$ is the only parameter variable in Equation \ref{eq:fai}. We can give the proof of Theorem \ref{thm:main} by proofing that equations \ref{eq:q1} and \ref{eq:q2} hold if and only if cross entropy $I\big(f_{\theta^*}^G(r(G,q^{G}_{\varphi^{*}}(G))),Y\big)$ reaches the maximum value.

First, we prove the sufficiency. According to the definition of mutual information, we have:
\begin{gather}
I\left(f_{\theta^*}^G\left(r\left(G,q^{G}_{\varphi^{*}}(G)\right)\right),Y\right) = \nonumber\\ H(Y) - H\left(Y|f_{\theta^*}^G\left(r\left(G,q^{G}_{\varphi^{*}}(G)\right)\right)\right).
\label{eq:I}
\end{gather}
$H(Y)$ is defined by the distribution of the dataset, therefore $H(Y)$ has a fixed value. Thus, if $I\left(f_{\theta^*}^G\left(r\left(G,q^{G}_{\varphi^{*}}(G)\right)\right),Y\right)$ reaches maximum, then $H\left(Y|f_{\theta^*}^G\left(r\left(G,q^{G}_{\varphi^{*}}(G)\right)\right)\right)$ reaches minimum, i.e., $H\left(Y|f_{\theta^*}^G\left(r\left(G,q^{G}_{\varphi^{*}}(G)\right)\right)\right) = 0$. Therefore, according to the definition of conditional entropy, we have:
\begin{gather}
H\left(Y|f_{\theta^*}^G\left(r\left(G,q^{G}_{\varphi^{*}}(G)\right)\right)\right)  \nonumber \\ = -\mathbb{E}\Big[log\Big( p\left(Y|f_{\theta^*}^G\left(r\left(G,q^{G}_{\varphi^{*}}(G)\right)\right)\right)\Big)\Big] = 0
\end{gather} 
As $log\Big( p\left(Y_{i}|f_{\theta^*}^G\left(r\left(G_{i},q^{G}_{\varphi^{*}}(G_{i})\right)\right)\right)\Big) \leq 0$, if the above equation holds, then:
\begin{gather}
log\Big( p\left(Y_{i}|f_{\theta^*}^G\left(r\left(G_{i},q^{G}_{\varphi^{*}}(G_{i})\right)\right)\right)\Big) = 0 \nonumber\\ , i = 1,2,3,...,n.
\end{gather} 
Therefore, we have:
\begin{gather}
p\left(Y_{i}|f_{\theta^*}^G\left(r\left(G_{i},q^{G}_{\varphi^{*}}(G_{i})\right)\right)\right) = 1 \nonumber\\ , i = 1,2,3,...,n.
\end{gather} 
As $f(\cdot)$ output the prediction of $Y$, thus, we have:
\begin{gather}
Y_i= f_{\theta^*}^G\left(r\left(G_i,q^{G}_{\varphi^{*}}(G_i)\right)\right) , i = 1,2,3,...,n.
\end{gather}
According to the assumptions, the following equation holds:
\begin{gather}
Y_{i}=f_{\theta^*}^G(X_{i}), i=1,2,3,...,n.
\end{gather}
And, $f^{G}_{\theta^*}(X_{i}\cup C'_{i})\neq Y_{i}, \forall C'_{i} \subseteq C_{i}$, therefore:
\begin{gather}
r\left(G_i,q^{G}_{\varphi^{*}}(G_i)\right)=X_{i}, i=1,2,3,...,n.
\label{eq:x=r}
\end{gather}
As $r(\cdot)$ is a non-parameter function that removes graph data according to $q(\cdot)$, according to Equation \ref{eq:x=r}, the confounder $C_{i}$ has been totally removed and the information of $X_{i}$ is unaffected. Therefore, the conditions defined in Equations \ref{eq:q1} and \ref{eq:q2} are met. Then, we can get that if $I\left(f_{\theta^*}^G\left(r\left(G,q^{G}_{\varphi^{*}}(G)\right)\right),Y\right)$ is maximized, then Equations \ref{eq:q1} and \ref{eq:q2} holds, the sufficiency is proved.

Next, we will prove the necessity. If Equations \ref{eq:q1} and \ref{eq:q2} hold, then $r(\cdot)$ will be able to remove $C_{i}$ without affecting $X_{i}$, therefore Equation \ref{eq:x=r} holds. Substituting Equation \ref{eq:x=r} into Equation \ref{eq:I}, we get:
\begin{gather}
I\left(f_{\theta^*}^G\left(r\left(G,q^{G}_{\varphi^{*}}(G)\right)\right),Y\right) = \nonumber \\I\left(f_{\theta^*}^G\left(X_{i}\right),Y\right) =  H(Y) - H\left(Y|f_{\theta^*}^G\left(X_{i}\right)\right).
\label{eq:Ire}
\end{gather}
As $Y_{i}=f^G_{\theta^*}(X_{i})$, we have:
\begin{gather}
I\left(f_{\theta^*}^G\left(r\left(G,q^{G}_{\varphi^{*}}(G)\right)\right),Y\right) = H(Y) - H(Y|Y).
\label{eq:Ire}
\end{gather}
Because $H(Y|Y) = 0$, therefore $I\left(f_{\theta^*}^G\left(r\left(G,q^{G}_{\varphi^{*}}(G)\right)\right),Y\right)$ is maximized. We can then come up with the conclusion that if Equation \ref{eq:q1} and Equation \ref{eq:q2} holds, then $I\left(f_{\theta^*}^G\left(r\left(G,q^{G}_{\varphi^{*}}(G)\right)\right),Y\right)$ is maximized. The necessity is proved.

With the sufficiency and necessity proved, we can say that Equations \ref{eq:q1} and \ref{eq:q2} holds if and only if cross entropy $H\big(f_{\theta^*}^G(r(G,q^{G}_{\varphi^{*}}(G))),Y\big)$ reaches the minimum value. According to the discussion at the beginning of the proof, we have proved Theorem \ref{thm:main}.

With Equation \ref{eq:goal1} and Theorem \ref{thm:main}, we define our IV approach-based training objective with unconditional moment restrictions as:
\begin{gather}
	 \theta^{*} = \arg\min_{\theta}\Big(\mathbb{E}[m(G_{i}, Y_{i}, q_{\varphi^{*}}^{G}(G_{i}), \theta)]\Big),
	\label{eq:goal2}
\end{gather}
where $\varphi^{*}$ can be calculated with Equation \ref{eq:fai}. However, at the beginning, we can neither acquire $\theta^{*}$ nor $\varphi^{*}$. Furthermore, the assumptions of Theorem \ref{thm:main} cannot be strictly met in real-world scenarios. To address such issues, we initialize $\theta$ and $\varphi$ randomly and update them alternatively. We further introduce some new designs to ensure that the model can learn label-causal-related graph representations against confounding effects. These designs will be discussed in the implementation and optimization parts.



\subsubsection{Implementation.}
We implement the framework to reach the training objectives defined in Equations \ref{eq:fai} and \ref{eq:goal2}. See Figure \ref{fig:framework} for the illustration. We build a GNN $f^{G}(\cdot)$ as our encoder, and a GNN $q^{G}(\cdot)$, which outputs the IVs as a series of edge weights. These edge weights are used to remove a certain amount of graph data, including dropping certain edges when the corresponding edge weights are too low and reducing information transfer through edges according to the edge weights. 

Meanwhile, some nodes and edges may contain both causal and confounding information. Certain critical information may be lost if they are removed before went through the GNN encoder. Therefore, we perform graph data removal after the $u$-th layer of $f^{G}(\cdot)$, $u$ is a hyperparameter. In this way, We can perform confounder elimination at a more fine-grained level.   

To make $f^{G}(\cdot)$ further robust to confounders, we introduce the concept of contrastive learning into our network. We enforce the encoder without the confounder removed to output similar features as the original ones (with the confounder removed), thereby bootstrapping the encoder's ability to eliminate the confounding effects of some confounders.



\subsubsection{Optimization.}


For optimization, we perform the updates alternatively, i.e., fix $\theta$ to learn $\varphi$, and then fix $\varphi$ to learn $\theta$, and so on. We update the parameter $\varphi$ of $q^{G}(\cdot)$ according to Theorem \ref{thm:main}, and maximize $I\left(f_{\theta^{*}}^G\left(r\left(G,q^{G}_{\varphi}(G)\right)\right), Y\right)$ in Equation \ref{eq:fai} by minimizing the cross entropy loss based on the label and the model output.

Our confounder elimination operation should be beneficial to improve the causality of model predictions. Therefore, in order to get a more ideal $\varphi$, we need to further optimize the training of $q^G(\cdot)$ from the causality aspect. According to Definition \ref{def:aug}, a confounder-robust GNN model is required to make correct predictions regardless of the confounders. Based on this, the model should learn the same representations for the \textit{same} and \textit{correct} category. For the graph sample $G_{i}$ of class $l$, suppose the model learns a representation vector $\boldsymbol{h}_{i}$, and $\boldsymbol{h}_{i}$ enables the model to perform the correct prediction. Then, the optimal value of $\boldsymbol{h}_{i}$ is to be the same as other representation vectors within the same class $l$. On the other hand, if $\boldsymbol{h}_{i}$ fails to enable the model to predict correctly, $\boldsymbol{h}_{i}$ is restricted to be different from all other classification feature vectors within class $l$. According to \cite{sun2020circle}, the farther the output deviates from the optimal value, the greater the contribution to training. Inspired by such intuition, during the optimization of $q^G(\cdot)$, we want to emphasize those samples that contribute more to training in terms of causality.

Specifically, for $\boldsymbol{h}_{i}$ belongs to class $l$, we first obtain a feature vector $\boldsymbol{t}_{i}$, which is the average of all feature vectors within the class $l$. Then, we acquire a series of weights $w_{i}$ for samples to perform the emphasizing operation, which is formulated by:
\begin{gather}
	 w_{i}= \begin{split}
 \left \{
\begin{array}{lr}
    |s(\boldsymbol{h}_{i}, \boldsymbol{t}_{i})-O_{max}|, & \text{correct result,}\\
    |s(\boldsymbol{h}_{i}, \boldsymbol{t}_{i})-O_{min}|, & \text{wrong result,}\\
\end{array}
\right.
\end{split}
	\label{eq:sim}
\end{gather}
where $s(\cdot)$ stands for the similarity calculation function, $O_{min}$ and $O_{max}$ denote the minimal and maximal values of similarity, respectively.

We can then define the loss for learning $\varphi$:

\begin{gather}
	 \mathcal{L}_{w} = \frac{1}{n}\sum_{i=1}^{n} w_{i}^{\gamma}\mathcal{H}\Big( f^G_{\dot{\theta}}\big(r(G_{i}, q^G_{\varphi}(G_{i}))\big), Y_{i}\Big),
	\label{eq:w}
\end{gather}
where $\mathcal{H}(\cdot)$ calculates the cross entropy loss for the $i$-th sample. $n$ is the amount of samples. $\dot{\theta}$ denotes the fixed parameters of $f^{G}(\cdot)$. $\gamma$ is a hyperparameter that controls the effect of $w_{l,i}$.

$\mathcal{L}_{w}$ makes the model output more consistent representations for the same class. We further hope that the representations of all the samples be discriminative and that the model can distinguish the samples of different classes. Therefore, we design a regular term $\mathcal{M}$:

\begin{gather}
	 \mathcal{M} = \tau \frac{1}{n} \sum_{i=1}^{n} |s(\boldsymbol{h}_{i}, \bar{\boldsymbol{t}})-O_{max}|,
	\label{eq:m}
\end{gather}

where $\bar{\boldsymbol{t}}$ denotes the average of all feature vectors of all classes. $\tau$ is a hyperparameter. $\mathcal{M}$ prevents the model from outputting the same features for all classes, or, falling into a local optimum as parts of the graph data are removed according to the proposed IV-based approach.

Leveraging Equations \ref{eq:w} and \ref{eq:m}, we propose the robustness-emphasizing loss for learning $\varphi$:

\begin{gather}
	 \mathcal{L}_{r} = \mathcal{L}_{w} + \zeta \mathcal{M},
	\label{eq:lr}
\end{gather}
where $\zeta$ can be defined as follows:


\begin{gather}
	 \zeta = \begin{split}
 \left \{
\begin{array}{lr}
    -1, & \mathcal{M} \leq \sigma\\
    0, & \mathcal{M} > \sigma.\\
\end{array}
\right.
\end{split}
	\label{eq:sim}
\end{gather}
$\sigma$ is the hyperparameter that determines when the regular term $\mathcal{M}$ should be activated.



For the optimization of $\theta$, our objective is defined in Equation \ref{eq:goal2}. As an implementation for $m(\cdot)$ in Equation \ref{eq:goal2}, we adopt the cross entropy loss to judge the difference between model output and ground truth labels. Formally, we define the loss for learning $\theta$ as:

\begin{gather}
	 \mathcal{L}_{o} = \frac{1}{n}\sum_{i=1}^{n} \mathcal{H}\Big( f^G_{\theta}\big(r(G_{i}, q^G_{\dot{\varphi}}(G_{i}))\big), Y_{i}\Big),
	\label{eq:lo}
\end{gather}
where $\dot{\varphi}$ denotes the fixed parameters of $q^G(\cdot)$. Moreover, we build a contrastive structure to make $f^G(\cdot)$ further robust to confounders. For the training objective, we design a contrastive loss as follows:

\begin{gather}
	 \mathcal{L}_{a} = -\frac{1}{n}\sum_{i=1}^{n}  s\Big(f^G_{\dot{\theta}}\big(r(G_{i}, q^G_{\dot{\varphi}}(G_{i}))\big), f'^G_{\theta}\big(G_{i}\big)\Big),
	\label{eq:la}
\end{gather}
where $s(\cdot)$ denotes the similarity calculation function, $\dot{\varphi}$ denotes the fixed parameters of $q^G(\cdot)$, and $f'^G(\cdot)$ denotes the encoder without the confounder removal module. 

Note that $f'^G(\cdot)$ and $f^G(\cdot)$ share the same GNN structure and parameters, but the parameters of $f^G(\cdot)$ are fixed as $\dot{\theta}$, i.e., $f^G(\cdot)$ are excluded from back propagation of $\mathcal{L}_{a}$. Therefore, $f'^G(\cdot)$ that fed by the input data containing confounders is enforced to generate similar outputs as $f^G(\cdot)$, which fed by the graphs without confounders (by using the confounder removal operation). Then, the learned parameters of $f'^G(\cdot)$ will be used to update $f^G(\cdot)$. This training paradigm encourages $f'^G(\cdot)$ to ignore information with confounding effects, result in certain confounders no longer have confounding effects on our model. Because we fix the outputs of a view in the contrastive learning part, we could avoid the risk of generating trivial solutions. Therefore, the contrastive learning in our method can be performed without negative samples, which significantly simplifies the computation during training.

We combine Equations \ref{eq:lo} and \ref{eq:la} to introduce the contrast-auxiliary loss as the training objective of $\theta$:
\begin{gather}
	 \mathcal{L}_{c} = \mathcal{L}_{o} + \lambda\mathcal{L}_{a},
	\label{eq:lc}
\end{gather}
where $\lambda$, as a hyperparameter, controls the effect of $\mathcal{L}_{a}$.

\section{Experiments}

\begin{table*}[ht]\small
	\setlength{\tabcolsep}{10pt}
	\begin{center}
		\begin{tabular}{l|cccc|c}
			\hline\rule{0pt}{10pt}

			\multirow{2}*{Method}  & \multicolumn{4}{c|}{Spurious-Motif}   &  Graph-SST2    \\
			\cline{2-5}\rule{0pt}{10pt}
			& Balanced & bias = 0.5 & bias = 0.7 & bias = 0.9 & (OOD) \\ 	
			\hline\rule{-3pt}{10pt}
			\text{ERM} & 42.99$\pm$1.93 & 39.69$\pm$1.73 & 38.93$\pm$1.74 & 33.61$\pm$1.02 & 81.44$\pm$0.59 \\
			\hline\rule{-3pt}{10pt}
			\text{GAT \cite{velivckovic2017graph}} & 43.07$\pm$2.55 & 39.42$\pm$1.50 & 37.41$\pm$0.86 & 33.46$\pm$0.43 & 81.57$\pm$0.71 \\
			\text{Top-k Pool \cite{gao2019graph}} &  43.43$\pm$8.79 & 41.21$\pm$7.05 & 40.27$\pm$7.12 & 33.60$\pm$0.91 & 79.78$\pm$1.35  \\
			\text{Group DRO \cite{sagawa2019distributionally}} &   41.51$\pm$1.11 & 39.38$\pm$0.93 & 39.32$\pm$2.23 & 33.90$\pm$0.52 &  81.29$\pm$1.44  \\

			\text{IRM \cite{arjovsky2019invariant}}  & 42.26$\pm$2.69 &  41.30$\pm$1.28 &  40.16$\pm$1.74 &  35.12$\pm$2.71 &   81.01$\pm$1.13 \\
			\text{V-REx \cite{krueger2021out}} & 42.83$\pm$1.59 &  39.43$\pm$2.69 &  39.08$\pm$1.56 &  34.81$\pm$2.04 &  81.76$\pm$0.08 \\
			\text{DIR \cite{wu2022discovering}}   & 42.53$\pm$3.38 &  41.45$\pm$2.12 &  41.03$\pm$1.53 &  39.20$\pm$1.94 &   81.93$\pm$1.26   \\
			\hline\rule{-3pt}{10pt}
			RCGRL-Wabl  & 43.49$\pm$2.06 &  42.97$\pm$1.96 &  42.06$\pm$1.53 &  40.83$\pm$1.89 &   81.93$\pm$0.89      \\
			RCGRL-Labl  & 43.84$\pm$1.98 &  43.41$\pm$1.24 &  41.94$\pm$1.39 &  40.63$\pm$1.52 &   81.64$\pm$1.13      \\
			\textbf{RCGRL}  & \bf{44.03$\pm$1.98} &  \bf{43.89$\pm$1.51} &  \bf{42.13$\pm$1.64} &  \bf{41.58$\pm$1.12} &   \bf{82.31$\pm$1.01}      \\
			\hline
		\end{tabular}
	\end{center}
	\caption{Performance of classification accuracy in OOD datasets, including Spurious-Motif and Graph-SST2. The Graph-SST2 dataset is artificially added with bias. We highlight the best records in bold.}
	\label{tab:b}
\end{table*}

\begin{table*}[ht]\small
	\setlength{\tabcolsep}{10pt}
	\begin{center}
		\begin{tabular}{l|ccccc}
			\hline\rule{0pt}{10pt}

			\multirow{2}*{Method}  &  Graph-SST2  & \multirow{2}*{Graph-Twitter} & \multirow{2}*{Mol-BBBP} & \multirow{2}*{Mol-BACE}   \\
			& (ID)   \\ 	
			\hline\rule{-3pt}{10pt}
			\text{ERM} & 89.19$\pm$0.87 & 65.29$\pm$1.05 & 64.19$\pm$0.80 & 73.61$\pm$0.75\\
			\hline\rule{-3pt}{10pt}
			\text{GAT \cite{velivckovic2017graph}} & 89.89$\pm$0.68 & 65.17$\pm$1.21 & 65.63$\pm$0.74 & 72.91$\pm$0.93\\
			\text{Top-k Pool \cite{gao2019graph}} & 89.31$\pm$1.26 & 63.78$\pm$0.97 & 64.69$\pm$1.41 & 71.30$\pm$0.84 \\
			\text{Group DRO \cite{sagawa2019distributionally}} & 89.94$\pm$1.60 & 63.35$\pm$0.52 &  64.35$\pm$1.26 & 70.38$\pm$1.47 \\

			\text{IRM \cite{arjovsky2019invariant}}  &  89.55$\pm$1.03 &  63.02$\pm$0.77 &   64.03$\pm$1.08 & 71.52$\pm$0.98\\
			\text{V-REx \cite{krueger2021out}} &  88.78$\pm$0.82 &  64.81$\pm$1.31 &  63.38$\pm$1.06 & 71.85$\pm$0.89\\
			\text{DIR \cite{wu2022discovering}}   &  89.91$\pm$0.86 &  65.14$\pm$0.96 &   65.36$\pm$1.14 & 72.30$\pm$1.39 \\
			\hline\rule{-3pt}{10pt}
			RCGRL-Wabl  & 89.95$\pm$0.47 &  66.36$\pm$0.38 &  66.02$\pm$0.99 &  74.03$\pm$0.61      \\
			RCGRL-Labl  & 90.38$\pm$0.75 &  \bf{66.73$\pm$0.49} &  65.69$\pm$1.17 &  73.31$\pm$0.83       \\
			\textbf{RCGRL}  &  \bf{90.73$\pm$0.40} &  66.58$\pm$0.53 &   \bf{66.78$\pm$0.95} & \bf{74.37$\pm$0.53}       \\
			\hline
		\end{tabular}
	\end{center}
	\caption{Performance in ID datasets, including classification accuracy in Graph-SST2 and Graph-Twitter, and ROC-AUC in Mol-BBBP and Mol-BACE. The best records are highlighted in bold.}
	\label{tab:nb}
\end{table*}

\begin{figure*}[ht]
	\centering
    \includegraphics[width=0.9\textwidth]{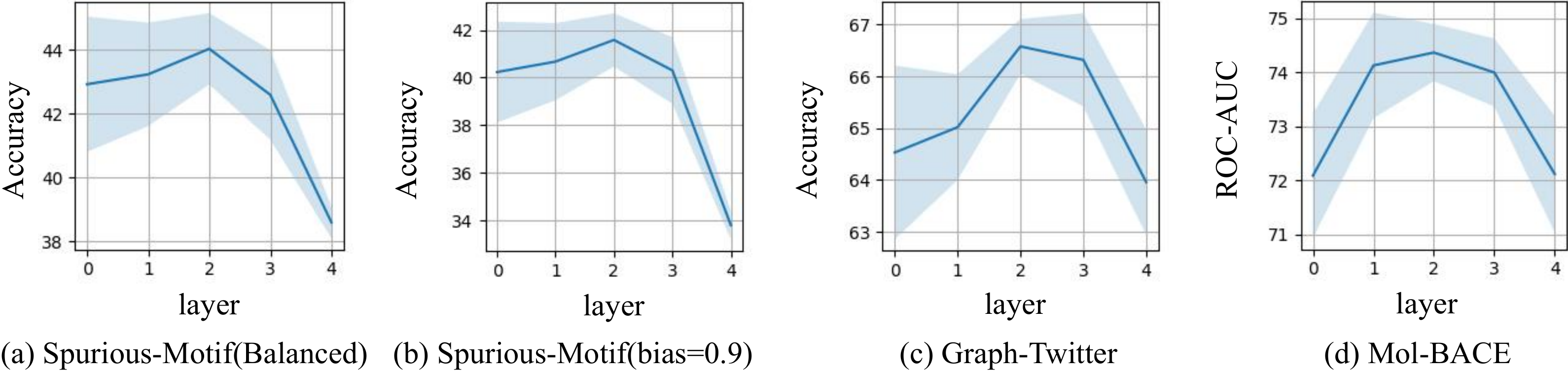}
	\caption{Performance of RCGRL with different $u$. If $u$ = 0, we introduce the IVs at the beginning. All models consist of four GNN layers.}
	\label{fig:chart}
\end{figure*}

\begin{figure*}[ht]
	\centering
	\subfigure[Confounder ratio]{
		\begin{minipage}{0.33\textwidth} 
			\includegraphics[width=\textwidth]{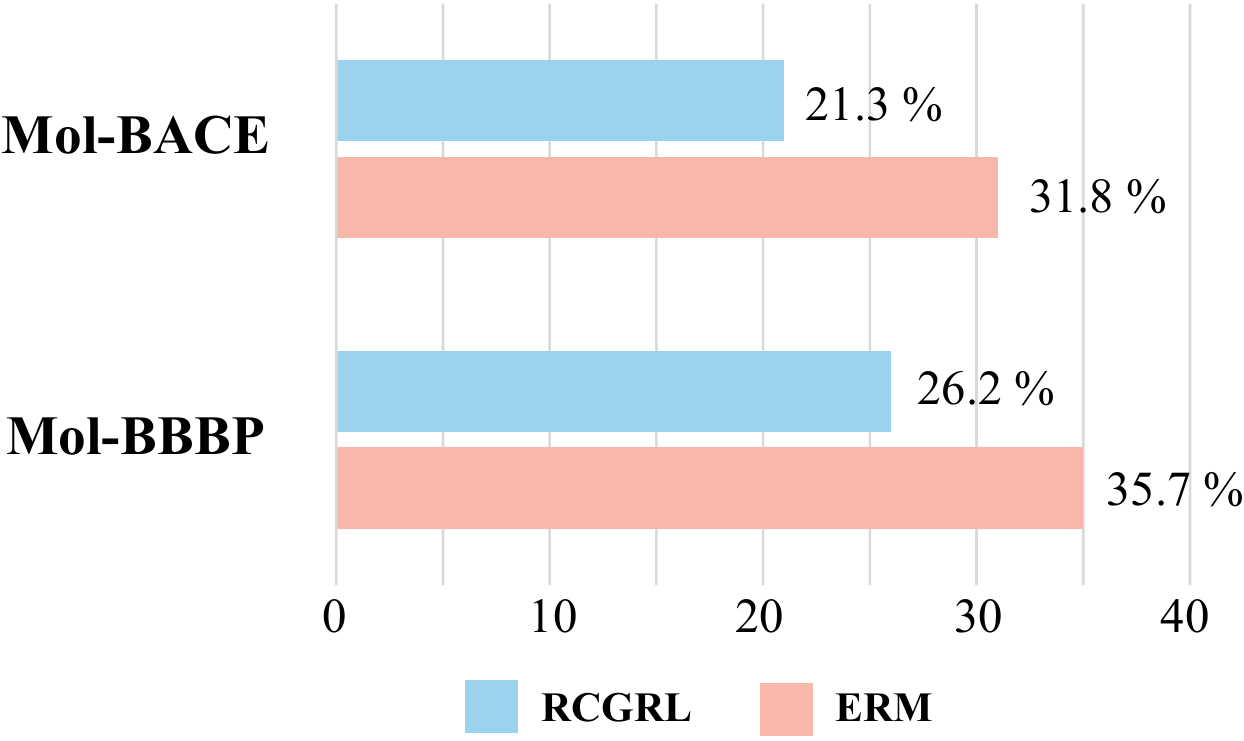} \\
            \label{fig:visa}
		\end{minipage}
	}
	\subfigure[Visualized results]{
		\begin{minipage}{0.45\textwidth}
			\includegraphics[width=\textwidth]{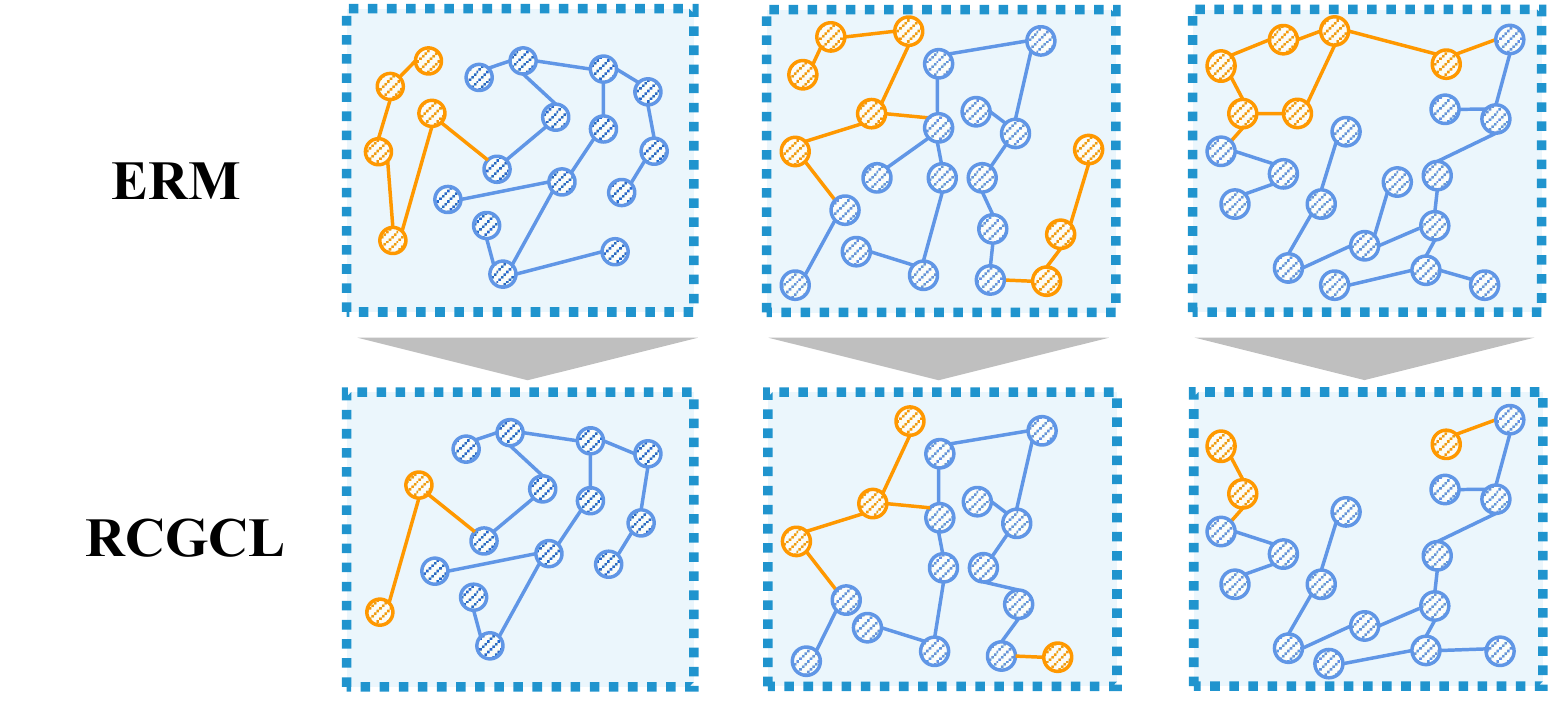} \\
            \label{fig:visb}
		\end{minipage}
	}
	
	\caption{The analyses of confounders. Visualized results show the confounders of several graphs in the datasets, and the orange substructures indicate the confounders measured by using the principle of Graph Granger Causality \cite{DBLP:conf/icml/LinLL21}.}
	\label{fig:vis}
\end{figure*}


\subsection{Comparison with State-of-the-art Methods }

\subsubsection{Datasets.} We evaluate our method on both OOD and ID datasets. The OOD datasets include: 1) Spurious-Motif, a synthetic dataset created by \cite{ying2019gnnexplainer}, and we adopt the re-implementation version created by \cite{wu2022discovering}; 2) Graph-SST2(OOD) is an OOD version of Graph-SST2 \cite{yuan2020explainability} created by \cite{wu2022discovering}. The ID datasets include: Graph-SST2 (ID) \cite{yuan2020explainability}, Graph-Twitter \cite{yuan2020explainability}, Mol-BBBP, and Mol-BACE \cite{hu2020open}. Different GNNs are picked for different datasets. Please refer to Appendix \ref{app:datasetsum} for details.

\subsubsection{Experiment Settings.} We compare our method with Empirical Risk Minimization (ERM) and various causality-enhanced methods, including the interpretable baselines, i.e., GAT and Top-k Pool, and the robust learning baselines, i.e., Group DRO, IRM, and DIR.

For a fair comparison, we follow the experimental principle of \cite{wu2022discovering} and adopt the same training setting for all models, which is described in detail in Appendix \ref{app:datasetsum}. For each task, we report the averaged performance $\pm$ std over ten runs. Some of the results are cited from \cite{wu2022discovering}. For the values of the hyperparameters, please refer to Appendix \ref{app:hyp}. To study the effects of the weights $w$ and contrastive loss $\mathcal{L}_{a}$, we create two additional ablation models by dropping certain structures. The first one is RCGRL-Wabl, which removes the weight factor $w$. The second one is RCGRL-Labl, which removes the contrastive loss $\mathcal{L}_{a}$. We train RCGRL-Wabl and RCGRL-Labl under the same training setting. 

\subsubsection{Results.}
The results are reported in Table \ref{tab:b} and \ref{tab:nb}. We observe from the tables and find that our method outperforms all baselines on all downstream tasks, which demonstrates that RCGRL can learn discriminative and robust graph representations against confounding effects. Such representations have the desired performance on both ID and OOD problems, i.e., RCGRL has outstanding prediction performance and generalization ability. For the ablation studies, RCGRL outperforms RCGRL-Wabl and RCGRL-Labl on most tasks, which verifies the effectiveness of our proposed weights $w$ and auxiliary contrastive learning.

\subsection{Evaluation of IV Introducing Position}
We introduce the IVs in the $u$-th layer instead of the beginning. Here, we provide an empirical analysis to prove the effectiveness of such a design. We conduct experiments on multiple datasets, including Spurious-Motif, Graph-Twitter, and Mol-BACE, with different $u$. As Figure \ref{fig:chart} shows, RCGRL reaches the best performance when $u = 2$ across all tasks. The reason behind this phenomenon is that introducing IV in the $u$-th layer helps the model to remove the confounders at the feature level, thereby avoiding excessive loss of semantic information of the original graphs.

\subsection{Evaluation of Confounder Robustness}
We conduct experiments on Mol-BACE and Mol-BBBP to demonstrate the confounder robustness of RCGRL. We follow \cite{DBLP:conf/icml/LinLL21} to measure the causality between the elements of the graph data and the ground-truth label. The elements that have no causal relationship with the label and may even affect the model's performance are considered confounders. We calculate the confounder percentage of graph representations learned by a trained ERM, which is compared with the confounder percentage of graph representations achieved by RCGRL. In Figure \ref{fig:visa}, the graphs with IV have a significant decrease in the proportion of confounders. Figure \ref{fig:visb} shows the visualized confounding effect information of several graphs, demonstrating that our method can remove a large number of confounders without causing a considerable loss of semantic information.

\section{Conclusion}

We propose RCGRL to learn robust graph representations against confounding effects. RCGRL actively generates instrumental variables under unconditional moment restrictions to eliminate confounders. Theoretically and empirically, we demonstrate the effectiveness of the proposed RCGRL.

\section*{Acknowledgements}
The authors would like to thank the anonymous reviewers for their valuable comments. This work is supported by National Key Research and Development Program of China No. 2019YFB1405100, CAS Project for Young Scientists in Basic Research, Grant No. YSBR-040.

\bibliography{aaai23}

\clearpage
\appendix
\begin{table*}[ht] \small
	\centering
	\begin{tabular}{lccccccc}
		\toprule
		Name      &  Graphs\# & Average Nodes\# & Classes\# & Task Type &Metric       \\
		\midrule
		Spurious-Motif &18,000 &46.6 &3 &Classification &ACC \\
		Graph-SST2 &70,042 &10.2 &2 &Classification &ACC \\
		Graph-Twitter &6,940 &21.1 &3 &Classification &ACC \\
		Mol-BACE &4,200 &27.0  &2 &Binary Classification &ROC-AUC \\
		Mol-BBBP &2,039 &24.1  &2 &Binary Classification &ROC-AUC \\
		\bottomrule
		
	\end{tabular}
	\caption{Summary of datasets.}
	\label{tab:datasum}
\end{table*}

\begin{table*}[ht] \small
	\centering
	\begin{tabular}{lccccc}
		\toprule
		Dataset  & Backbone & Size of $f^G$ & Size of $q^G$ (GNN) & Size of $q^G$ (MLP) & Global Pool       \\
		\midrule
		Spurious-Motif &Local Extremum GNN &[4,32,32,32] &[4,32] &[64,1] & global mean pool  \\
		Graph-SST2 &ARMA &[768,128,128,2] &[768,128] &[256,1] & global mean pool  \\
		Graph-Twitter &ARMA &[768,128,128,2] &[768,128] &[256,1] & global mean pool  \\
		Mol-BACE &GIN + Virtual nodes &[9,300,300,300,1] &[9,300] &[600,1] &global add pool \\
		Mol-BBBP &GIN + Virtual nodes &[9,300,300,300,1] &[9,300] &[600,1] &global add pool \\
		\bottomrule
	\end{tabular}
	\caption{Summary of the backbones used in each dataset.}
	\label{tab:network}
\end{table*} 

\section{Appendix: Experiment Setting Details} \label{app:datasetsum}
In this section, we give the details of our experiment settings. 

\subsection{Datasets}
We evaluate our method on five different datasets, including manually constructed datasets, sentiment graph datasets, and molecule datasets. We summarize the datasets we used in Table \ref{tab:datasum}. We also describe the backbone GNN we adopt for each dataset in Table \ref{tab:network}, including Local Extremum GNN \cite{DBLP:conf/aaai/RanjanST20}, ARMA \cite{DBLP:conf/aaai/0001RFHLRG19}, GIN \cite{xu2018powerful}, and Virtual nodes \cite{DBLP:conf/nips/HuFRNDL21}. Some of the datasets have artificially added confounders.

(1) Spurious-Motif dataset. Spurious-Motif is a synthetic dataset created by \cite{ying2019gnnexplainer}. We adopt the re-implementation version created by \cite{wu2022discovering}, which involves 18, 000 graphs. Each graph in the dataset consists of two subgraphs. One serves as the ground-truth data that is causally related to the graph label, the other serves as the confounder. There are three types of ground truth and confounder subgraphs each. In the training set, the confounder subgraph was stitched to the graph data with a certain bias. e.g., if the bias is 0.7, that means for each class, 70$\%$ of the samples are stitched with the same kind of confounder subgraph. The ground-truth subgraphs and confounder subgraphs in the test sets are randomly attached. Therefore, the higher the bias, the harder for the model to distinguish the causal information from confounders. Additionally, the dataset includes graphs with large subgraphs to make the learning task harder.

(2) Graph-SST2 dataset (OOD) \cite{yuan2020explainability}. Graph-SST2 is a sentiment graph dataset. We follow \cite{wu2022discovering} to split the graph samples into different sets according to their average node degree to increase the difficulty of the training.


(3) Graph-SST2 dataset (ID) \cite{yuan2020explainability}. The original Graph-SST2 dataset.

(4) Graph-Twitter dataset \cite{yuan2020explainability}. Also, a sentiment graph dataset as Graph-SST2 dataset, but with different data sources.

(5) Mol-BBBP \& Mol-BACE \cite{hu2020open}. Molecule datasets from OGBG datasets.

\subsection{Environments and Optimization Method}
All our experiments were conducted on a workstation with two Quadro RTX 5000 GPU (16 GB), one Intel Xeon E5-1650 CPU, 128GB RAM, and an Ubuntu 20.04 operating system. During training, we adopt the Adam optimizer. We set the maximum training epoch as 300 for all tasks. For backpropagation, We use Stochastic Gradient Descent (SGD) for the optimization on Graph-SST2, Graph-Twitter, Mol-BBBP, and Mol-BACE, and Gradient Descent (GD) for the Spurious-Motif.

\subsection{Hyperparameters and Implementation Details}
\label{app:hyp}

As for hyperparameters, we set $u$, the hyperparameter that controls the IV introducing position as 2, i.e., introducing IV at layer 2. For the robustness-emphasizing loss, we set $w$ as 0.1 for Spurious-Motif dataset, 0.08 for Mol-BACE and Mol-BBBP datasets, and 0.05 for Graph-SST2 and Graph-Twitter datasets. We set $\sigma$, $\lambda$, and $\tau$ as 0.01, 1.0 and 1000 for all datasets. For learning rate, we set the learning rate at 0.001 for Spurious-Motif dataset, and 0.0002 for the others. We train each task for 30 epochs, after which we stop the training early if there exist 5 epochs without performance improvement on the validation set. We pick the model with the best validation performance as the final result. The max training epochs number is set as 300 for all datasets. The training batch size of Spurious-Motif, Graph-SST2, and Graph-Twitter datasets is 32. For Mol-BACE and Mol-BBBP, the batch size is 256.

For the implementation of $q(\cdot)$, we adopt a GNN to generate representation vectors for the nodes. The structure of the GNNs are shown in Table \ref{tab:network}. Then, we utilize a two-layer mlp to generate the edge weight of each edge based on the representation vectors of the attached vertexes of that edge. After that, we remove a certain number of edges (75$\%$ for Spurious-Motif, 40$\%$ for Graph-SST2 and Graph-Twitter, 20$\%$ for Mol-BACE, and Mol-BBBP) based on the edge weights.

\end{document}